\documentclass{article} 
\usepackage{iclr2026_conference,times}

\usepackage{amsmath,amsfonts,bm}

\def\eqref#1{equation~\ref{#1}}

\def\1{\bm{1}}

\DeclareMathAlphabet{\mathsfit}{\encodingdefault}{\sfdefault}{m}{sl}
\SetMathAlphabet{\mathsfit}{bold}{\encodingdefault}{\sfdefault}{bx}{n}

\usepackage{amsthm}

\usepackage{hyperref}
\usepackage{url}
\usepackage{caption}
\usepackage[most]{tcolorbox}
\usepackage{tabularx}
\usepackage[most]{tcolorbox}
\usepackage{xspace}
\usepackage{booktabs}   
\usepackage{multirow}   
\usepackage{makecell}   
\usepackage[ruled,vlined]{algorithm2e}
\SetKwInput{KwInput}{Input}
\SetKwInput{KwOutput}{Output}
\usepackage[table]{xcolor} 
\definecolor{grey}{rgb}{0.5,0.5,0.5}
\usepackage{wrapfig}   

\usepackage{graphicx}
\usepackage{float}
\usepackage{amsmath}
\usepackage{amssymb}
\usepackage{enumitem}
\usepackage{listings}
\usepackage{subcaption}

\lstdefinestyle{promptstyle}{
  basicstyle=\ttfamily\footnotesize,
  backgroundcolor=\color{gray!10},
  frame=single,
  breaklines=true,
  columns=fullflexible
}

\theoremstyle{definition}

\title{Towards Verifiable Agentic Data Science: \\ Solving Irregular TSQA via Tool-Grounded Reasoning}

\author{Sanhorn Chen, Xiaoyang Chen, Boyu Liu, Roy Zhao \\
University of Illinois Urbana Champaign\\
\{sanhorn2, xc52\}@illinois.edu\\}

\iclrfinalcopy 
\begin{document}

\maketitle
\fancyhead{}        
\renewcommand{\headrulewidth}{0pt} 

\begin{abstract}
Time series data in real-world deployments is overwhelmingly irregular. Observations are asynchronous, missing values are informative rather than random, and sampling frequencies vary across sensors and operational windows. However, existing Time Series Question Answering (TSQA) benchmarks mostly assume regularly sampled inputs, leaving a fundamental gap in understanding how large language models (LLMs) and AI agents perform under irregular conditions. To bridge this gap, we introduce \emph{IRTS-ToolBench}, a benchmark of 1,700 questions spanning 10 task types across 13 domains. \emph{IRTS-ToolBench} is designed to be used independently by any researcher working on LLM-based irregular time series analysis, providing standardized inputs and a reproducible evaluation protocol. Code can be found in \url{https://github.com/SanhornC/IRTS-ToolBench}.
\end{abstract}

\section{Introduction}
Time series data is among the most common forms of information in real-world systems~\citep{chang2025timeimmdatasetbenchmarkirregular}. 
The rapid development of Large Language Models (LLMs)   and AI agents   for complex reasoning tasks have made Time Series Question Answering (TSQA) an increasingly important evaluation paradigm. Recent benchmarks~\citep{jing2026tsaqatimeseriesanalysis, kong2025timemqatimeseriesmultitask, wu2026scitsscientifictimeseries, yu2026tsrbenchcomprehensivemultitaskmultimodal, yin2026mmtsbench} have demonstrated that LLMs can perform temporal reasoning over structured time series inputs with contextual information as supplement. Concurrently, agentic frameworks~\citep{wu2026timeartagentictimeseries, liu2026tsagentunderstandingreasoningraw, zhao2025timeseriesscientistgeneralpurposeaiagent} have shown that tool-augmented and multi-agents outperform direct LLM inference on complex time series tasks. However, these works share a critical limitation: they operate exclusively on regular time series while assumption that time series data arrives at regular is systematically violated in practice. For instance, ICU monitoring systems record vital signs when a nurse or physician deems it clinically necessary, and industrial sensors log readings only when anomalies are detected. As a result, the absence of a rigorous benchmark for irregular time series constitutes a fundamental gap in our evaluation infrastructure.

Existing synthetic irregularization methods, such as MCAR random dropout and sparse mask resampling, apply transformations without understanding the domain semantics or the reason behind the missingness, resulting in data that are statistically irregular but semantically implausible. TIME-IMM~\citep{chang2025timeimmdatasetbenchmarkirregular} identifies this gap, demonstrating that real-world irregularity is organized into nine cause-driven types across three categories (Trigger-Based, Constraint-Based, and Artifact-Based) none of which are captured by random dropout. To the best of our knowledge, no prior work has proposed pipelines for semantically grounded regular-to-irregular time series transformation.

We introduce \emph{IRTS-ToolBench}, a benchmark specifically designed to evaluate LLMs and AI agents on irregular univariate TSQA via tool-grounded reasoning. Our contributions are threefold. (1) We propose a novel LLM-Guided Irregular Transformation Pipeline for semantically grounded regular-to-irregular time series transformation. (2) The \emph{IRTS-ToolBench} consists of 1,700 questions spanning 10 task types across 13 domains. (3) We collected a 30-tool library with 7 irregularity operators and 23 analytical tools, enabling the benchmark to support evaluation of both LLMs and AI agents. 

The remainder of this paper is organized as follows. Sec. \ref{sec:related_works} situates IRTS-ToolBench within related work. Sec. \ref{sec:benchmark_design} describes the benchmark design in detail, including the transformation pipeline, task descriptions, and tool library. Sec. \ref{sec:eval} defines the evaluation protocol. Sec. \ref{sec:benchmark_validation} discusses benchmark validation. Sec. \ref{sec:conclusion} concludes with limitations and future directions.

\section{Related Work}
\label{sec:related_works}
\textbf{TSQA Benchmarks.} TSAQA~\citep{jing2026tsaqatimeseriesanalysis} provides a hierarchical taxonomy of six task types, including anomaly detection, classification, characterization, comparison, data transformation, and temporal relationship, across 60 datasets. Its multi-LLM consensus question generation mechanism directly inspired our own pipeline. Time-MQA~\citep{kong2025timemqatimeseriesmultitask} extends TSQA to multi-task settings with context enhancement. ITFormer~\citep{wang2025itformerbridgingtimeseries} introduces a aero-engine QA dataset pairing time series with natural language for cross-modal alignment tasks. Despite their contributions, all three benchmarks assume clean, regularly sampled time series as input. 

\textbf{Irregular Time Series: Taxonomy and Transformation Methods.} TIME-IMM~\citep{chang2025timeimmdatasetbenchmarkirregular} introduces a nine-type irregularity taxonomy organized into Trigger-Based, Constraint-Based, and Artifact-Based categories, each anchored to a real-world dataset, and critiques existing benchmarks for assuming regular sampling. Our transformation pipeline uses this taxonomy as its decision space. \citep{du2025pypotspythontoolkitmachine} provide implementations of MCAR, MAR, MNAR, block missing, and sequence missing functions, forming the execution layer of our pipeline. Physiome-ODE~\citep{klötergens2025physiomeodebenchmarkirregularlysampled} provides an irregular multivariate forecasting benchmark based on biological ODEs. Critically, none of these works employ LLMs to guide the irregularization process.

\textbf{Agentic Time Series Frameworks.} TimeART~\citep{wu2026timeartagentictimeseries} proposes a ReAct-style agent with curated tools for time series reasoning, establishing the template for tool-augmented agentic evaluation that our benchmark extends to the irregular domain. TS-Agent~\citep{liu2026tsagentunderstandingreasoningraw} focuses on statistical insight gathering for time series tasks. TimeSeriesScientist~\citep{zhao2025timeseriesscientistgeneralpurposeaiagent} proposed agent with different roles optimized for forecasting workflows.
\section{Benchmark Design}
\label{sec:benchmark_design}
\subsection{Task Domain, Scope, and Descriptions}
IRTS-ToolBench draws from two source datasets~\citep{jing2026tsaqatimeseriesanalysis, kong2025timemqatimeseriesmultitask} spanning 13 domains (e.g. finance, healthcare). All collected regular TS samples are reconstructed through a three-layer pipeline (Sec.~\ref{sec:benchmark_construction_pipeline}) into univariate irregular TS in one of \textit{multiple\_choices (MC)} or \textit{true\_or\_false (TF)} format.
The benchmark is organized into \emph{10} task types grouped into \emph{3} categories (Details of Tasks Descriptions in Table~\ref{tab:task_summarization}): (1) \textbf{Standard Reasoning}: This category of questions require models to perform simple time series reasoning primitives over irregular time series inputs and test whether models can identify differences between irregular and regular TS. Specifically, we introduce \emph{4} tasks: \emph{Anomaly Detection}, \emph{Classification}, \emph{Regular vs. Irregualr Discrimination}, and \emph{Regularity Recovery}. 
(2) \textbf{Irregularity-Specific Reasoning}: In this category, we evaluate whether models can reason over irregular TS inputs by identifying their temporal characteristics, inferring relationships among asynchronous observations, and attributing the underlying causes of irregular sampling patterns. Specifically, we introduce \emph{3} tasks: \emph{Characterization}, \emph{Temporal Relationship}, and \emph{Irregularity Cause Attribution}. (3) \textbf{Regularity–Irregularity Interface Reasoning}: These tasks require models to reason across the boundary between regular and irregular representations, and test whether models can reason at the interface between regularity and irregularity. Specifically, we introduce \emph{3} tasks \emph{Missingness Reasoning}, \emph{Irregularity Severity Estimation}, and \emph{Forecasting}. These three categories are designed to form a progressive reasoning hierarchy, moving from basic irregular sampling, to explicit understanding of irregular temporal structures, and finally to reasoning across the interface between regular and irregular representations. For more details regarding task descriptions, please refer to Appendix~\ref{sec:a_task_design}. Additionally, task examples can be found in Appendix~\ref{appendix:examples}. 

\subsection{Benchmark Construction Pipeline}
\label{sec:benchmark_construction_pipeline}
Our benchmark draws from 2 source datasets~\citep{jing2026tsaqatimeseriesanalysis, kong2025timemqatimeseriesmultitask}. Specifically, each collected sample is processed through a three-layer transformation pipeline. We describe each pipeline stage below.

\textbf{Irregular Time Series Transformation.}
Our transformation pipeline, visualized in Figure~\ref{fig:time_mqa} leverages the irregularity types from \citep{chang2025timeimmdatasetbenchmarkirregular} as a decision space for selecting the appropriate transformation mechanism for each sample. Specifically, the pipeline follows a unified three-stage design. (1) \textbf{Context Enrichment}:  We leverage a LLM to generate an enriched context description capturing the domain, statistical features, and signal characteristics for each time series sequence. (2) \textbf{Taxonomy Selection}: Given the enriched context and time series statistics, an LLM selects the most appropriate irregularity type and outputs a transformation plan with associated anomaly annotations along with a self-evaluated confidence score.
(3) \textbf{Parameter Generation}: An LLM translates the transformation plan into validated numerical parameters for the relevant transformation functions. Transformation execution then applies the generated parameters, producing an irregular time series with realistic missingness patterns and optionally jittered timestamps.  More information can be found in Appendix~\ref{sec:a_irregular_transform}.



\textbf{Question Generation Pipeline.} For each task type, a dedicated question generation pipeline produces questions conditioned on the transformed irregular time series and its associated anomaly annotations. The generation procedure, visualized in Figure~\ref{fig:qg}, is as follows:
(1) \textbf{Primary Generation}: GPT-5.1 is prompted with task-specific prompts to generate the \emph{meta\_information} for each sample. Depending on the task design, the corresponding \emph{question} (\textit{MC} or \textit{TF}) and answer are either generated by GPT-5.1 or constructed using task-specific deterministic functions. (2) \textbf{Multi-LLM Consensus}: Then, three independent LLMs (GPT-5.1, Claude Sonnet 4.5, Gemini 2.5 Flash) evaluate whether the generated \emph{meta\_information}, \emph{question}, and \emph{answer} are clear and answerable from the provided time series (e.g. score $>$ threshold). More details in Appendix~\ref{sec:qgp}.


\textbf{Golden Tool Set Construction.} For each finalized benchmark sample, a golden tool set specifies the minimum set of tools that a model must invoke to correctly answer the question. The construction protocol, visualized in Figure~\ref{fig:golden_tool_set}, proceeds as follows: 
(1) \textbf{Independent Proposal}: Three LLMs (GPT-5.1, Claude Sonnet 4.5, Gemini 2.5 Flash) independently propose the required tool sequence for the question, together with a self-confidence score and an explanation for each tool's necessity. (2) \textbf{Consensus Gold Set}: The Gold Set is constructed by \emph{Majority Voting} with a \emph{Union} fallback, which tools proposed by at least two LLMs are retained, while if no tool receives majority support, the union of all proposed tools is used. More details in Appendix~\ref{sec:a_golden}.

\subsubsection{Tool Library}
IRTS-ToolBench provides a 30-tool library, as shown in Table \ref{table:tool_library}, that enables the benchmark to support evaluation of both LLMs (via tool-augmented prompting) and AI agents (via agentic tool-use frameworks). The library is organized into two layers: (1) \textbf{Irregularity Operators}: These tools provide the operations for handling irregular time series sequences. (2) \textbf{Advanced Analytical Tools}: This layer consists of 23 tools, providing time series analysis primitives, such as summary statistics and trend and seasonality detection. More details in Appendix~\ref{sec:tool_library}.

\section{Evaluation Protocol}
\label{sec:eval}
\emph{IRTS-ToolBench} is designed to be evaluated under a standardized protocol. Specifically, the model output is parsed and compared with the ground-truth, both in either \textit{MC} or \textit{TF} question format.  We define three primary metrics and specify how they should be computed, aggregated, and reported.

\textbf{Metrics.} The primary metric is \emph{Overall Accuracy}: \text{Accuracy} = $\frac{1}{N}\sum_{i=1}^{N}s_i$, where N is the total number of evaluated examples. For both \textit{MC} and \textit{TF} questions, the model output must exactly match the ground-truth answer to receive credit. We also report \emph{Task-Level Accuracy} for each task type $t$: $\text{Accuracy}(t)=\frac{1}{|\mathcal{D}_t|}\sum_{i\in\mathcal{D}_t}s_i$, where $D_t$ is the set of examples belonging to task $t$. In addition, when tool evaluation is enabled, we compare the set of tools called by the model with the proposed golden tool set. Each example is assigned one label: (1) \emph{Exact Match}, where the called tool set matches the golden tool set; (2) \emph{Partial Match}, where the called tool set partially matches the golden tool set; (3) \emph{Complete Mismatch}, where the called tool set doesn't match the golden tool set. For each task type, we report these three rates. These rates are mutually exclusive and sum to 1 within each task, and they are not used to compute overall accuracy and task-level accuracy.

\textbf{Scoring and Aggregation.}
All questions are scored using binary scoring. Each example receives $1$ if the model answer matches the ground-truth, and $0$ otherwise. \emph{Overall accuracy} is the average score across all samples, while \emph{Task-Level Accuracy} is the average score within each task type. Tool-match rates are aggregated separately by task type and used as indicators of tool-use behavior.

\textbf{Reporting Requirements.}
For each model, we report \emph{Overall Accuracy}, \emph{Task-Level Accuracy} across the $10$ task types, and, when applicable, task-level \emph{Tool-Match Rates}. We also specify whether the model is evaluated in tool-augmented mode.

\section{Benchmark Validation}
\label{sec:benchmark_validation}
\textbf{Task Quality Verification.}
Task quality is controlled in both \emph{Question Generation} and \emph{Golden Tool Set Construction} pipeline through a multi-LLM evaluator layer. For \emph{Question Generation} pipeline, each generated QA pair is reviewed by GPT-5.1, Claude Sonnet 4.5, and Gemini 2.5 Flash using the same evaluation prompt. Evaluators assign confidence scores based on question clarity, answer support from the provided time series, and the validity of distractors. Most tasks are retained when all evaluators score greater or equal to a threshold. For instance, Anomaly Detection task requires all evaluators to score at least 0.85. For \emph{Golden Tool Set Construction} pipeline, given each QA pair and the proposed tool library, GPT-5.1, Claude Sonnet 4.5, and Gemini 2.5 Flash independently propose tools required to solve the question. A tool is collected if it appears in at least two model proposals. If no tool reaches majority support, we use the union of all proposed tools as a fallback. 

\textbf{Human Peer Review and Human Baseline.}
We further conduct human peer-review testing on a randomly sampled 2\% subset of the benchmark, stratified by task type. Two undergraduate reviewers independently answer the sampled questions and flag ambiguous or underspecified items. This evaluation serves two purposes: validating task clarity and providing a simple human baseline. The two reviewers achieve overall accuracies of 80\% and 78\%, respectively. Overall uncertainty rates are low, although \emph{Regularity Recovery} and \emph{Missingness Reasoning} show higher uncertainty and lower agreement than the other tasks. These results suggest that most benchmark questions are understandable to human reviewers, while the harder irregularity-specific tasks remain challenging.

\textbf{Evaluation Results.} We evaluate zero-shot performance of (1) commercial LLMs: Claude-Opus-4.7 (without thinking, with thinking, with tool-calling) (2) open-source LLMs: Qwen3.5-4B~\citep{qwen3.5}, Qwen3.6-27B~\citep{qwen3.6-27b}, and DeepSeek-V4-Flash~\citep{deepseekai2026deepseekv4} (with and without tool-calling). Our \emph{Overall Results}, as shown in Table~\ref{tab:eval_results}, suggest that Qwen3.6-27B (78.59) outperforms the other three models in all settings, while commerical model Claude-Opus-4.7 receive a consistent overall accuracy in between 74 to 77 percent. For \emph{Task-Level Results}, tool calling provides the most visible improvements on several tasks. For open-source models, Qwen3.6-27B improves from 96.80 to 99.60 on anomaly detection and reaches 100.00 on classification with tools, while DeepSeek-V4-Flash shows a particularly large gain on irregularity severity estimation task, increasing from 31.33 to 98.67, and also improves on regularity recovery from 64.67 to 89.33. On the other hand, Claude-Opus-4.7 is generally strong on irregularity cause attribution and missingness reasoning task. However, some tasks remain difficult even with tool use, especially temporal relationship reasoning and regular-vs-irregular discrimination, where performance is less consistent across models. Overall, these results suggest that for irregular time series QA, tool calling is beneficial for tasks requiring explicit numerical analysis, while higher-level temporal reasoning remains challenging. For more details, please refer to Appendix~\ref{sec:a_eval_results}.
\section{Conclusion}
\label{sec:conclusion}
We proposed \emph{IRTS-ToolBench}, a benchmark of 1,700 questions across 10 task types and 13 domains for evaluating LLMs and AI agents on irregular univariate time-series question answering. By combining semantically grounded irregular time-series construction, task-level diagnostic, and a 30-tool library with golden tool sets, \emph{IRTS-ToolBench} provides a standardized evaluation for studying both answer correctness and tool-grounded reasoning behavior. Our evaluation results show that current LLMs already exhibit non-trivial reasoning ability over irregular time series, especially when contextual information and tools are available, but they still struggle with higher-level temporal reasoning and require strong scaffolding for reliable agentic tool use. A key limitation is that our three-layer LLM-based generation pipeline may inherit sensitivity from LLM prompting, consensus quality, and transformation choices, so future work will strengthen pipeline robustness through broader validation and more naturally irregular source data. We also plan to extend the benchmark toward more complex multi-hop TSQA settings and incorporate additional modalities such as visual plots to better reflect real-world time-series analysis scenarios.


\bibliography{iclr2026_conference}

@misc{chang2025timeimmdatasetbenchmarkirregular,
      title={Time-IMM: A Dataset and Benchmark for Irregular Multimodal Multivariate Time Series}, 
      author={Ching Chang and Jeehyun Hwang and Yidan Shi and Haixin Wang and Wen-Chih Peng and Tien-Fu Chen and Wei Wang},
      year={2025},
      eprint={2506.10412},
      archivePrefix={arXiv},
      primaryClass={cs.LG},
      url={https://arxiv.org/abs/2506.10412}, 
}

@misc{jing2026tsaqatimeseriesanalysis,
      title={TSAQA: Time Series Analysis Question And Answering Benchmark}, 
      author={Baoyu Jing and Sanhorn Chen and Lecheng Zheng and Boyu Liu and Zihao Li and Jiaru Zou and Tianxin Wei and Zhining Liu and Zhichen Zeng and Ruizhong Qiu and Xiao Lin and Yuchen Yan and Dongqi Fu and Jingchao Ni and Jingrui He and Hanghang Tong},
      year={2026},
      eprint={2601.23204},
      archivePrefix={arXiv},
      primaryClass={cs.AI},
      url={https://arxiv.org/abs/2601.23204}, 
}

@misc{kong2025timemqatimeseriesmultitask,
      title={Time-MQA: Time Series Multi-Task Question Answering with Context Enhancement}, 
      author={Yaxuan Kong and Yiyuan Yang and Yoontae Hwang and Wenjie Du and Stefan Zohren and Zhangyang Wang and Ming Jin and Qingsong Wen},
      year={2025},
      eprint={2503.01875},
      archivePrefix={arXiv},
      primaryClass={cs.CL},
      url={https://arxiv.org/abs/2503.01875}, 
}

@misc{wu2026timeartagentictimeseries,
      title={TimeART: Towards Agentic Time Series Reasoning via Tool-Augmentation}, 
      author={Xingjian Wu and Junkai Lu and Zhengyu Li and Xiangfei Qiu and Jilin Hu and Chenjuan Guo and Christian S. Jensen and Bin Yang},
      year={2026},
      eprint={2601.13653},
      archivePrefix={arXiv},
      primaryClass={cs.LG},
      url={https://arxiv.org/abs/2601.13653}, 
}

@misc{liu2026tsagentunderstandingreasoningraw,
      title={TS-Agent: Understanding and Reasoning Over Raw Time Series via Iterative Insight Gathering}, 
      author={Penghang Liu and Elizabeth Fons and Annita Vapsi and Mohsen Ghassemi and Svitlana Vyetrenko and Daniel Borrajo and Vamsi K. Potluru and Manuela Veloso},
      year={2026},
      eprint={2510.07432},
      archivePrefix={arXiv},
      primaryClass={cs.AI},
      url={https://arxiv.org/abs/2510.07432}, 
}

@misc{zhao2025timeseriesscientistgeneralpurposeaiagent,
      title={TimeSeriesScientist: A General-Purpose AI Agent for Time Series Analysis}, 
      author={Haokun Zhao and Xiang Zhang and Jiaqi Wei and Yiwei Xu and Yuting He and Siqi Sun and Chenyu You},
      year={2025},
      eprint={2510.01538},
      archivePrefix={arXiv},
      primaryClass={cs.LG},
      url={https://arxiv.org/abs/2510.01538}, 
}

@misc{wang2025itformerbridgingtimeseries,
      title={ITFormer: Bridging Time Series and Natural Language for Multi-Modal QA with Large-Scale Multitask Dataset}, 
      author={Yilin Wang and Peixuan Lei and Jie Song and Yuzhe Hao and Tao Chen and Yuxuan Zhang and Lei Jia and Yuanxiang Li and Zhongyu Wei},
      year={2025},
      eprint={2506.20093},
      archivePrefix={arXiv},
      primaryClass={cs.CL},
      url={https://arxiv.org/abs/2506.20093}, 
}

@misc{du2025pypotspythontoolkitmachine,
      title={PyPOTS: A Python Toolkit for Machine Learning on Partially-Observed Time Series}, 
      author={Wenjie Du and Yiyuan Yang and Linglong Qian and Jun Wang and Qingsong Wen},
      year={2025},
      eprint={2305.18811},
      archivePrefix={arXiv},
      primaryClass={cs.LG},
      url={https://arxiv.org/abs/2305.18811}, 
}

@misc{klötergens2025physiomeodebenchmarkirregularlysampled,
      title={Physiome-ODE: A Benchmark for Irregularly Sampled Multivariate Time Series Forecasting Based on Biological ODEs}, 
      author={Christian Klötergens and Vijaya Krishna Yalavarthi and Randolf Scholz and Maximilian Stubbemann and Stefan Born and Lars Schmidt-Thieme},
      year={2025},
      eprint={2502.07489},
      archivePrefix={arXiv},
      primaryClass={cs.LG},
      url={https://arxiv.org/abs/2502.07489}, 
}

@misc{qwen3.6-27b,
    title  = {{Qwen3.6-27B}: Flagship-Level Coding in a {27B} Dense Model},
    author = {{Qwen Team}},
    month  = {April},
    year   = {2026},
    url    = {https://qwen.ai/blog?id=qwen3.6-27b}
}

@misc{deepseekai2026deepseekv4,
      title={DeepSeek-V4: Towards Highly Efficient Million-Token Context Intelligence},
      author={DeepSeek-AI},
      year={2026},
}

@misc{qwen3.5,
    title  = {{Qwen3.5}: Towards Native Multimodal Agents},
    author = {{Qwen Team}},
    month  = {February},
    year   = {2026},
    url    = {https://qwen.ai/blog?id=qwen3.5}
}

@misc{yu2026tsrbenchcomprehensivemultitaskmultimodal,
      title={TSRBench: A Comprehensive Multi-task Multi-modal Time Series Reasoning Benchmark for Generalist Models}, 
      author={Fangxu Yu and Xingang Guo and Lingzhi Yuan and Haoqiang Kang and Hongyu Zhao and Lianhui Qin and Furong Huang and Bin Hu and Tianyi Zhou},
      year={2026},
      eprint={2601.18744},
      archivePrefix={arXiv},
      primaryClass={cs.AI},
      url={https://arxiv.org/abs/2601.18744}, 
}

@misc{wu2026scitsscientifictimeseries,
      title={SciTS: Scientific Time Series Understanding and Generation with LLMs}, 
      author={Wen Wu and Ziyang Zhang and Liwei Liu and Xuenan Xu and Jimin Zhuang and Ke Fan and Qitan Lv and Junlin Liu and Chen Zhang and Zheqi Yuan and Siyuan Hou and Tianyi Lin and Kai Chen and Bowen Zhou and Chao Zhang},
      year={2026},
      eprint={2510.03255},
      archivePrefix={arXiv},
      primaryClass={cs.LG},
      url={https://arxiv.org/abs/2510.03255}, 
}

@misc{
yin2026mmtsbench,
title={{MMTS}-Bench: A Comprehensive Benchmark for Multimodal Time Series Understanding and Reasoning},
author={Yao Yin and Zhenyu Xiao and Musheng Li and Yiwen Liu and Sutong Nan and Yiting He and Ruiqi Wang and Zhenwei Zhang and Yuantao Gu},
year={2026},
url={https://openreview.net/forum?id=PMKpyXk0FO}
}
\bibliographystyle{iclr2026_conference}

\appendix
\newpage

\section{Task Design}
\label{sec:a_task_design}
\begin{table}[h]
\scriptsize
\centering
\begin{tabular}{c|l|p{4cm}|l|c}
\toprule
{\bf Group} & {\bf Task} & {\bf Description} & {\bf Question Type} & {\bf Number of Samples} \\
\midrule
\multirow{7}{*}{\centering\makecell[c]{Standard Reasoning}}
& Anomaly Detection & Tests whether the model can identify abnormal observations in irregular time series given contextual metadata. & MC & 250 \\
\cmidrule{2-5}
& Classification & Tests coarse-grained pattern recognition under irregular sampling conditions. There are several areas of questions that can be asked such as classifying the input's domain. & MC, TF & 150 \\
\cmidrule{2-5}
& \makecell[l]{Regular vs. Irregular \\ Discrimination} & Tests the model's ability to identify whether a given sequence is regularly or irregularly sampled. & MC, TF & 400 \\
\cmidrule{2-5}
& Regularity Recovery & Tests whether the model can identify the most plausible regular reconstruction of an irregular input. & MC & 150 \\
\midrule
\multirow{5}{*}{\centering\makecell[c]{Irregularity-Specific\\Reasoning}}
& Characterization & Tests the model's ability to describe temporal properties (trend, seasonality, stationarity) of an irregular sequence. & MC, TF & 100 \\
\cmidrule{2-5}
& Temporal Relationship & Tests reasoning about ordering, causality, and lag relationships across asynchronous observations. & MC & 150 \\
\cmidrule{2-5}
& \makecell[l]{Irregularity Cause \\ Attribution} & Tests whether the model can identify the underlying sampling mechanism (Trigger-Based / Artifact-Based / Constraint-Based) of a given irregular sequence. & MC & 100 \\
\midrule
\multirow{5}{*}{\centering\makecell[c]{Regularity–Irregularity \\ Interface Reasoning}}
& Missingness Reasoning & Tests whether the model can distinguish informative gaps from non-informative gaps in an irregular sequence. & MC & 100 \\
\cmidrule{2-5}
& \makecell[l]{Irregularity Severity \\ Estimation} & Tests whether the model can determine the interval of irregularity severity for a given sequence. & MC & 150 \\
\cmidrule{2-5}
& Forecasting / Imputation & Tests forward-looking prediction and missing-value estimation under irregular conditions. & MC & 150 \\
\bottomrule
\end{tabular}
\caption{Task Summarization. TF and MC denote true\_or\_false and multiple\_choices.}
\label{tab:task_summarization}
\end{table}

\subsection{Task Descriptions}
We describe all $10$ tasks organized in three categories as shown in Table \ref{tab:task_summarization}. For each task we specify the input format, expected output, success criteria, and the primary challenge it poses for models and agents.

\subsubsection{Standard Reasoning}
\textbf{Anomaly Detection.} Model is given an irregular univariate time series together with domain metadata and anomaly annotations. The question asks whether a specific anomaly exists, where it occurs, or how many anomalies are present. This task is formulated as \textit{MC} question type. This task is challenging because anomaly evidence may be embedded in sparse, non-uniformly spaced observations, requiring models to reason about the absolute timestamps, and apply appropriate missingness handling before pattern analysis.

\textbf{Classification.} Given an irregular univariate sequence with meta context, the model must classify which domain the input time series comes from a given set of concurrent choices. The primary challenge is that standard classification heuristics are not directly applicable to irregular input. Models must adapt analytical reasoning to non-uniform observation spacing or invoke appropriate resampling tools before classification. Questions are formulated as either \textit{MC} or \textit{TF} format. 

\textbf{Regular vs. Irregular Discrimination.} The model is given a time series, either a regular sequence or an irregular sequence transformed by our three-layer pipeline, and asked to determine whether it is regularly or irregularly sampled. For \textit{MC} questions, the model selects from options such as 'regular (uniform spacing)', 'irregular (random missingness)', 'irregular (event-driven)', or 'irregular (jittered timestamps)'. For \textit{TF} questions, the model judges a single claim about the sampling regularity. This task is challenging because even a visually irregular-looking sequence may exhibit near-regular spacing after mild jitter, requiring models to compute inter-observation interval statistics rather than relying on surface-level appearance.

\textbf{Regularity Recovery.} The model is given an irregular time series and asked to select the most plausible regular reconstruction from \textit{MC} options that differ in resampling frequency, interpolation method, or imputed value ranges. Ground truth is derived from the original regular sequence prior to transformation. This task requires models to first identify the appropriate regular grid, then determine which interpolation or imputation strategy best recovers the original signal, and finally evaluate candidate reconstructions against this criterion. 

\subsection{Irregularity-Specific Reasoning}

\textbf{Characterization.} The model is asked to describe temporal properties of an irregular sequence, including trend direction (upward / downward / flat), seasonality presence and approximate period, and stationarity status. Questions are either \textit{MC} or \textit{TF}. This task is challenging because, in the irregular setting, models must either apply interpolation or use tools specifically designed for irregular sequence characterization before making a judgment.

\textbf{Irregularity Cause Attribution.} The model is given an irregular time series together with a domain description and asked to identify which of the nine irregularity category~\citep{chang2025timeimmdatasetbenchmarkirregular} best explains the observed missingness pattern. Questions are \textit{MC}. Ground truth is the irregularity type assigned by our transformation pipeline. This task is novel and challenging because distinguishing irregularity requires integrating domain knowledge with structural pattern analysis.

\textbf{Temporal Relationship.} This task evaluates the model’s ability to infer the temporal structure among irregular time series patches. This task is formulated as \textit{MC} question format. Given the first chronological patch $\mathbf{x}$, an \textit{MC} question asks the model to choose the correct next patch from candidates $[\mathbf{y}_1,\mathbf{y}_2,\mathbf{y}_3,\mathbf{y}_4]$.
The false candidates are randomly sampled from the full dataset, but from sequences different from that of $\mathbf{x}$.

\subsubsection{Regularity–Irregularity Interface Reasoning}

\textbf{Missingness Reasoning.} The model is given an irregular time series and asked to distinguish whether a specific gap or pattern of missing values constitutes an informative signal or a non-informative gap. Questions are formulated as \textit{MC} format. The ground truth is derived from the type of irregularity and the domain context used in the transformation. This task is challenging because models must invoke tools to reason about gap structure before rendering a judgment.

\textbf{Irregularity Severity Estimation.} The model is given an irregular time series and asked to estimate which observation rate interval describes the sequence's missingness density. Question is formulated as \textit{MC} format. Ground truth is the observation rate computed during transformation. This task directly tests a model's perception of missingness density. It is challenging because models without explicit tool invocation must estimate missingness density from the raw sequence representation alone, which requires counting observed and missing entries across potentially long sequences without making positional errors.

\textbf{Forecasting.} The model is given an irregular time series up to a specified observation boundary and asked to predict the value or range of the target variable at a future timestamp. The question is presented in \textit{MC} format with quantized answer buckets to enable exact-match evaluation. Ground truth is derived from the original regular series. This task requires models to reason about the series' trend and local dynamics while accounting for the irregular observation spacing leading up to the forecast horizon.

\newpage
\section{Benchmark Construction Pipeline}
Our benchmark draws from 2 source datasets~\citep{jing2026tsaqatimeseriesanalysis, kong2025timemqatimeseriesmultitask}. Specifically, each collected sample is processed through a three-layer transformation pipeline in three dimensions: (1) All time series sequences are transformed from regular into irregular format; (2) All questions are synthetically generated based on the transformed sequences and their anomaly annotations; and (3) Golden tool sets are constructed via a separate multi-LLM consensus protocol.

\subsection{Irregular Time Series Transformation}
\label{sec:a_irregular_transform}

\begin{figure}[h]
    \centering
    \includegraphics[width=1\linewidth]{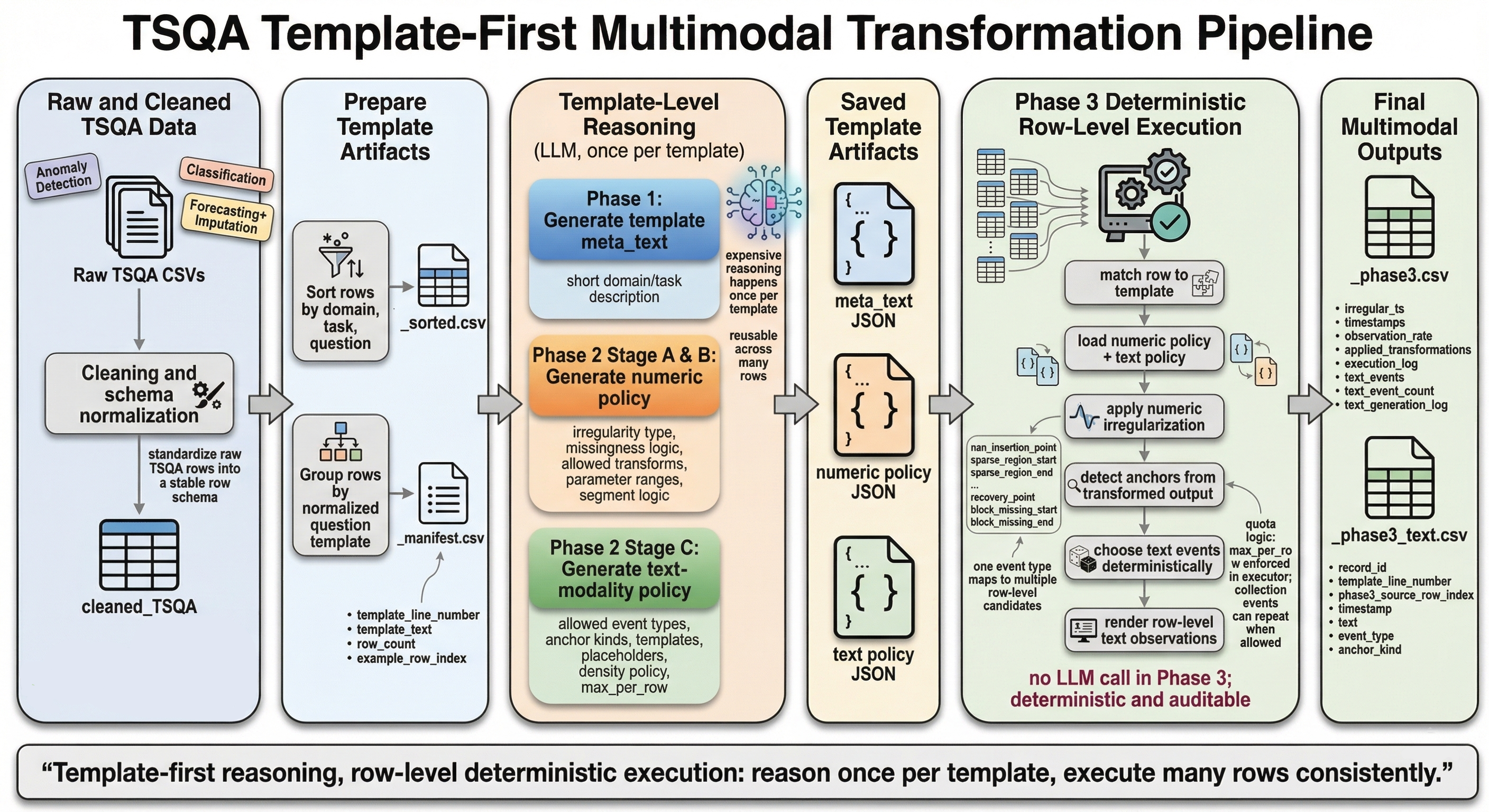}
    \caption{Irregular Time Series Transformation.}
    \label{fig:time_mqa}
\end{figure}

Our transformation pipeline leverages the irregularity types from TIME-IMM~\citep{chang2025timeimmdatasetbenchmarkirregular} as a principled decision space for selecting the appropriate transformation mechanism for each sample. Specifically, the pipeline follows a unified three-stage design as shown in figure~\ref{fig:time_mqa}. (1) \textbf{Context Enrichment}:  We leverage a LLM to generate an enriched context description capturing the domain, statistical features, and signal characteristics for each time series sequence. (2) \textbf{Taxonomy Selection}: Given the enriched context and time series statistics, an LLM selects the most appropriate irregularity type and outputs a transformation plan with associated anomaly annotations along with a self-evaluated confidence score.
(3) \textbf{Parameter Generation}: An LLM translates the transformation plan into validated numerical parameters for the relevant transformation functions. Transformation execution then applies the generated parameters, producing an irregular time series with realistic missingness patterns and optionally jittered timestamps. 

\subsection{Question Generation Pipeline}
\label{sec:qgp}
\begin{figure}[t]
    \centering
    \includegraphics[width=1\linewidth]{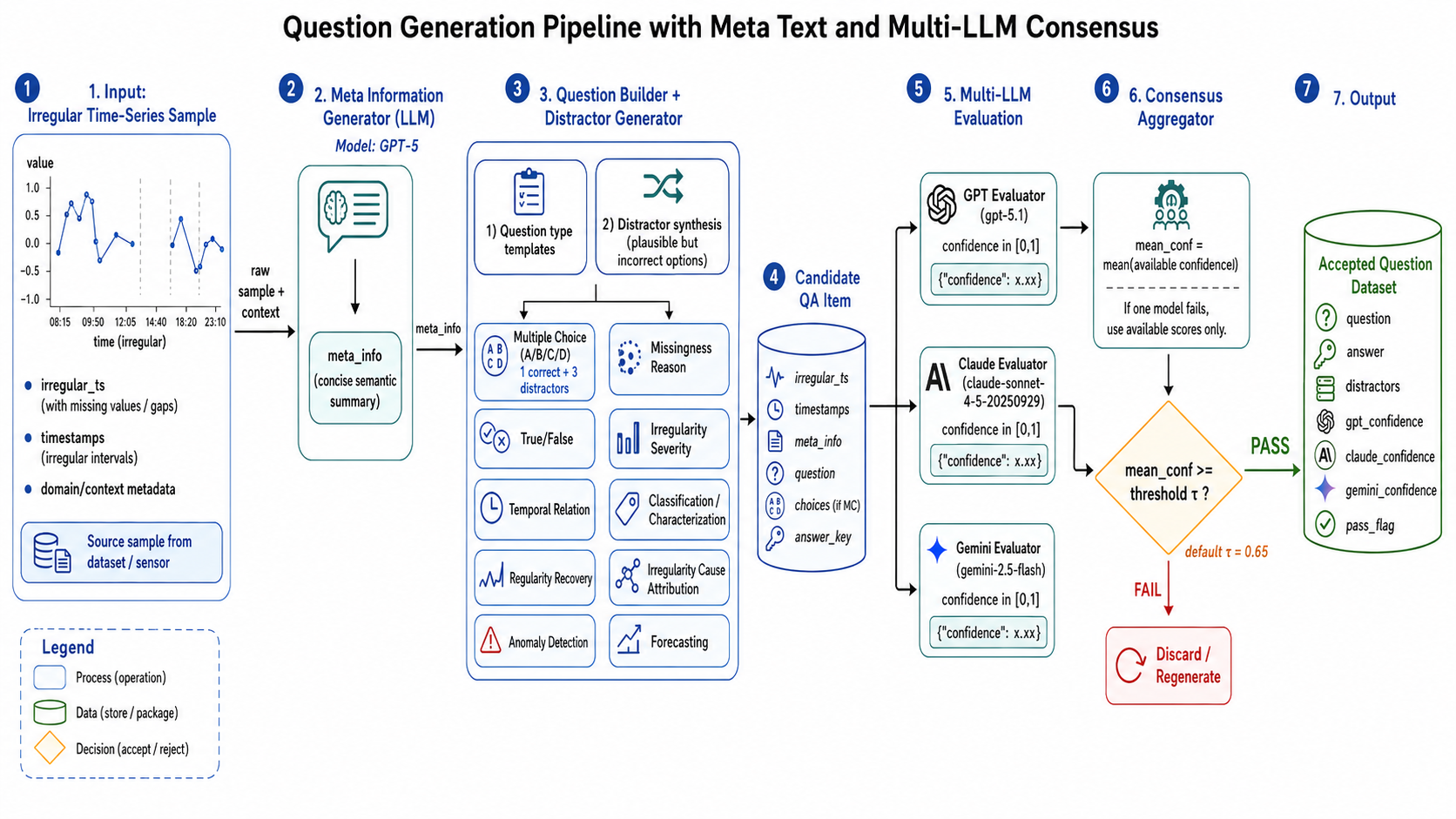}
    \caption{Question Generation Pipeline.}
    \label{fig:qg}
\end{figure}

For each task type, a dedicated question generation pipeline, as shown in Figure~\ref{fig:qg}, produces questions conditioned on the transformed irregular time series and its associated anomaly annotations. The generation procedure is as follows:
(1) \textbf{Primary Generation}: GPT-5.1 is prompted with task-specific prompts to generate the \emph{meta\_information} for each sample. Depending on the task design, the corresponding \emph{question} (\textit{MC} or \textit{TF}) and answer are either generated by GPT-5.1 or constructed using task-specific deterministic functions. (2) \textbf{Multi-LLM Consensus}: Then, three independent LLMs (GPT-5.1, Claude Sonnet 4.5, Gemini 2.5 Flash) evaluate whether the generated \emph{meta\_information}, \emph{question}, and \emph{answer} are clear and answerable from the provided time series (e.g. score $>$ threshold).

\subsection{Golden Tool Set Construction}
\label{sec:a_golden}
\begin{figure}[h]
    \centering
    \includegraphics[width=1\linewidth]{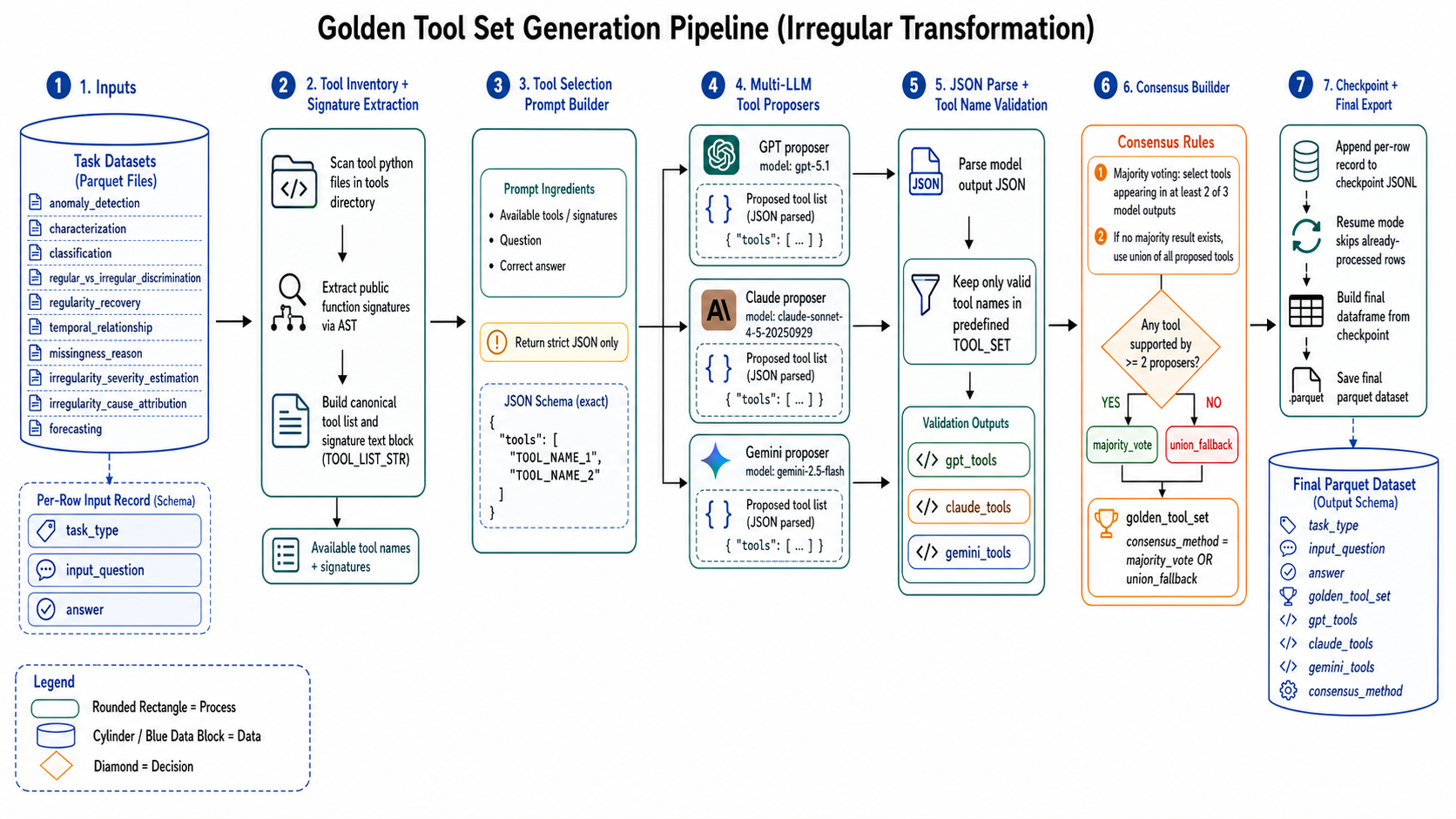}
    \caption{Golden Tool Set Construction.}
    \label{fig:golden_tool_set}
\end{figure}

For each finalized benchmark sample, a golden tool set, as shown in Figure~\ref{fig:golden_tool_set}, specifies the minimum set of tools that a model must invoke to correctly answer the question. The construction protocol proceeds as follows: 
(1) \textbf{Independent Proposal}: Three LLMs (GPT-5.1, Claude Sonnet 4.5, Gemini 2.5 Flash) independently propose the required tool sequence for the question, together with a self-confidence score and an explanation for each tool's necessity. (2) \textbf{Consensus Gold Set}: The Gold Set is constructed by \emph{Majority Voting} with a \emph{Union} fallback, which tools proposed by at least two LLMs are retained, while if no tool receives majority support, the union of all proposed tools is used.

\section{Evaluation Results}
\label{sec:a_eval_results}
We evaluate zero-shot performance of (1) commercial LLMs: Claude-Opus-4.7 (without thinking, with thinking, with tool-calling) (2) open-source LLMs: Qwen3.5-4B~\citep{qwen3.5}, Qwen3.6-27B~\citep{qwen3.6-27b}, and DeepSeek-V4-Flash~\citep{deepseekai2026deepseekv4} (with and without tool-calling). 

\textbf{Overall Results.} Our evaluation result, as shown in Table~\ref{tab:eval_results} suggests that Qwen3.6-27B (78.59) outperforms the other three models in all settings, while commerical model Claude-Opus-4.7 receive a consistent overall accuracy in between 74 to 77 percent. 

\textbf{Task-Level Results.} For standard tasks such as anomaly detection and classification, tool calling leads to clear gains for several models. In particular, Qwen3.6-27B improves from 96.80 to 99.60 on anomaly detection and reaches 100.00 on classification with tool use, while Claude-Opus-4.7 also achieves strong performance on both tasks. However, smaller open-source models remain less stable without tools, especially on anomaly detection, where Qwen3.5-4B obtains only 42.40 accuracy. This suggests that irregular time-series questions still require reliable numerical reasoning even for seemingly conventional task types. 

Tasks that require deeper interpretation of irregular patterns show larger performance gaps across models. Claude-Opus-4.7 generally perform strongly on irregularity cause attribution and missingness reasoning. In contrast, open-source models show more uneven behavior: Qwen3.5-4B performs poorly on irregularity severity estimation, while larger models such as Qwen3.6-27B and DeepSeek-V4-Flash benefit substantially from tool calling. For example, DeepSeek-V4-Flash improves from 31.33 to 98.67 on ISE and from 64.67 to 89.33 on RR when tools are enabled.

\begin{table*}[h]
    \centering
    \scriptsize
    \setlength{\tabcolsep}{3.0pt}
    \begin{tabular}{c|l|c|cccccccccc|c}
        \toprule
        \textbf{Type} & \textbf{Model} & \textbf{Tool} 
        & \textbf{A.D.} & \textbf{Char.} & \textbf{CLS} & \textbf{FCA} 
        & \textbf{ICA} & \textbf{ISE} & \textbf{MR} & \textbf{RvI} 
        & \textbf{RR} & \textbf{TR} & \textbf{Overall} \\
        \midrule

        \multirow{3}{*}{Commercial}
        & Claude-Opus-4.7 (w/o thinking) & No & 88.40 & 94.50 & 40.00 & -- & 99.00 & 84.67 & 91.00 & -- & 37.33  & \textbf{62.00} & 74.88  \\
        & Claude-Opus-4.7 (thinking)     & No & 87.60 & 95.00 & 38.00 & -- & 97.00 & 95.33 & 91.00 & -- & 49.33 & 58.67 & 76.72 \\
        & Claude-Opus-4.7 (thinking)     & Yes & 86.40 & \textbf{98.00} & \textbf{100.00} & \textbf{64.67} & \textbf{100.00} & 79.33 & \textbf{92.00} & 79.25 & 74.67 & 52.00 & 76.00 \\
        
        \midrule
        
        \multirow{7}{*}{\makecell{Open-\\source}}
        & \multicolumn{13}{l}{\textit{Without tool calling}} \\
        \cmidrule(lr){2-14}
        & Qwen3.5-4B & No 
        & 42.40 & 93.00 & 98.67 & 35.33 
        & 92.00 & 10.00 & 36.00 & 67.25 
        & 38.00 & 46.00 & 55.18 \\
        
        & Qwen3.6-27B & No 
        & 96.80 & 97.00 & 99.33 & 56.00 
        & \textbf{100.00} & 58.00 & 91.00 & \textbf{81.75} 
        & 58.00 & 48.00 & \textbf{78.59} \\
        
        & DeepSeek-V4-Flash & No 
        & 59.20 & 92.00 & 90.67 & 51.33
        & 84.00 & 31.33 & 66.00 & 52.00 
        & 64.67 & 46.67 & 60.29 \\
        
        \cmidrule(lr){2-14}
        & \multicolumn{13}{l}{\textit{With tool calling}} \\
        \cmidrule(lr){2-14}
        & Qwen3.6-27B & Yes 
        & \textbf{99.60} & 94.00 & \textbf{100.00} & 46.67 
        & 96.00 & 81.33 & 88.00 & 61.50 
        & 60.00 & 40.00 & 74.41 \\
        
        & DeepSeek-V4-Flash & Yes 
        & 96.40 & 79.00 & 92.00 & 47.33 
        & 86.00 & \textbf{98.67} & 80.00 & 58.00 
        & \textbf{89.33} & 32.67 & 74.00 \\
        
        \bottomrule
    \end{tabular}
    \caption{Evaluation results on IRTS-ToolBench. Tool indicates whether tool calling is enabled. A.D. denotes anomaly detection. CLS denotes classification. FCA denotes forecasting. ICA denotes irregularity cause attribution. ISE denotes irregularity severity estimation. MR denotes missingness reasoning. RvI denotes regular-vs-irregular discrimination. RR denotes regularity recovery. TR denotes temporal relationship.}
    \label{tab:eval_results}
\end{table*}
\section{Tool Library}
\label{sec:tool_library}
IRTS-ToolBench provides a 30-tool library, shown in Table \ref{table:tool_library} that enables the benchmark to support evaluation of both LLMs (via tool-augmented prompting) and AI agents (via agentic tool-use frameworks). The library is organized into two layers: (1) \textbf{Irregularity Operators}: These tools provide the operations for handling irregular time series sequences. (2) \textbf{Advanced Analytical Tools}: This layer consists of 23 tools, providing time series analysis primitives, such as summary statistics and trend and seasonality detection. The full tool list with descriptions is shown in Table \ref{table:tool_library}.

\begin{table}[t]
\centering
\tiny
\resizebox{\textwidth}{!}{
    \begin{tabular}{l|p{8cm}}
    \toprule
    \textbf{Tool Name} & \textbf{Description} \\ 
    \midrule
    \multicolumn{2}{c}{\textbf{Irregularity Operators}} \\
    \midrule
    ALIGN\_ASOF & Asynchronous-of join: aligns features observed at different timestamps onto a common time axis \\ 
    \midrule
    MISSING\_POLICY & Determines and applies an imputation or masking strategy based on missingness type (MCAR / MAR / MNAR) \\ 
    \midrule
    RESAMPLE & Resamples irregular timestamps onto a regular grid at a specified frequency \\ 
    \midrule
    IRREGULARITY\_SCORE & Computes an irregularity degree measure for a sequence (gap variance, observation rate, timestamp CV) \\ 
    \midrule
    GAP\_DETECT & Identifies gap segments in a time series and characterizes their length distribution \\ 
    \midrule
    EVENT\_MAP & Maps irregular observations to known event timestamps for event-driven alignment \\ 
    \midrule
    TIMESTAMP\_JITTER\_CORRECT & Estimates and corrects timestamp offsets caused by scheduling jitter or clock drift \\ 
    \midrule
    \multicolumn{2}{c}{\textbf{Advanced Analytical Tools}} \\
    \midrule
    SERIES\_INFO & Returns metadata: sequence length, channel count, missing value statistics \\ 
    \midrule
    DATAPOINT\_VALUE & Returns all channel values at a specified time index or timestamp \\ 
    \midrule
    SUMMARY\_STATS & Computes mean, sum, max, min, std over a specified index range \\ 
    \midrule
    ROLLING\_STAT & Computes rolling statistics (mean, sum, max, min, std) with a sliding window \\ 
    \midrule
    QUANTILE\_VALUE & Returns empirical quantile values (e.g., q=0.5 for median) per channel \\ 
    \midrule
    VOLATILITY & Computes rolling volatility (std of first differences) over a given window \\ 
    \midrule
    AUTOCORR & Computes autocorrelation coefficient at a specified lag per channel \\ 
    \midrule
    TREND\_CLASSIFIER & Classifies trend as up / down / flat; supports global and window-based segment analysis \\ 
    \midrule
    SEASONALITY\_DETECTOR & Detects periodic patterns and returns estimated period with strength indicator \\ 
    \midrule
    CHANGE\_POINT\_DETECTOR & Detects structural breaks in mean or variance and returns change point indices \\ 
    \midrule
    STATIONARITY\_TEST & Tests stationarity via ADF or KPSS; returns status and test statistics \\ 
    \midrule
    SPIKE\_DETECTOR & Detects spikes or dips using amplitude threshold and minimum separation \\ 
    \midrule
    NOISE\_PROFILE & Labels noise type (white / red) based on autocorrelation tests \\ 
    \midrule
    CHANNEL\_CORRELATION & Computes Pearson or Spearman correlation between two channels with optional lag \\ 
    \midrule
    CROSS\_CORRELATION & Computes cross-correlation across multiple lags to find optimal time alignment \\ 
    \midrule
    DTW\_DISTANCE & Measures similarity between two channels using Dynamic Time Warping \\ 
    \midrule
    SHAPE\_SIMILARITY & Measures shape similarity between channels, invariant to amplitude scaling \\ 
    \midrule
    GRANGER\_CAUSALITY & Tests statistical predictability between channels within a specified lag \\ 
    \midrule
    ANOMALY\_DETECTION & Zero-shot anomaly detection via reconstruction-error-based model (DADA) \\ 
    \midrule
    FORECASTER & Zero-shot multivariate forecasting via lightweight foundation model (LightGTS) \\ 
    \midrule
    INTERPOLATE & Performs linear, spline, or forward-fill imputation for missing value handling \\ 
    \midrule
    DIFF\_TRANSFORM & Computes first-order differencing for stationarity analysis or data transformation tasks \\ 
    \midrule
    WAVELET\_TRANSFORM & Applies wavelet decomposition for multi-scale characterization tasks \\ 
    \bottomrule
    \end{tabular}
}
\caption{Full list of tools of Tool Library.}
\label{table:tool_library}
\end{table}
\newpage
\section{Examples}
\label{appendix:examples}

In this section, we select one representative task example from each reasoning level.
For Standard Reasoning, we select \emph{Regular vs. Irregular Discrimination} task as example. For Irregularity-Specific Reasoning, we select \emph{Irregularity Cause Attribution} task as example. For Regularity-Irregularity Interface Reasoning, we select \emph{Missingness Reasoning} task as example. Each example will consist of three inputs: 
(1) meta\_info (optional depending on task), (2) irregular time series sequence, and (3) question.

\begin{tcolorbox}[
  enhanced, breakable, title= IRTS-ToolBench — Regular vs. Irregular Discrimination, colback=grey, colframe=black!70, boxrule=0.8pt,
  colbacktitle=black!85, coltitle=white, fonttitle=\bfseries, title filled,
  sharp corners=all, arc=1.5mm, left=2mm, right=2mm, top=2mm, bottom=2mm
]
\begin{tcolorbox}[
  enhanced, breakable, title=Time Series Info ,
  colback=black!1, colframe=black!20, boxrule=0.5pt, title filled=false
]
\begin{minipage}[t]{1\linewidth}
  \includegraphics[height=6cm, width=\linewidth]{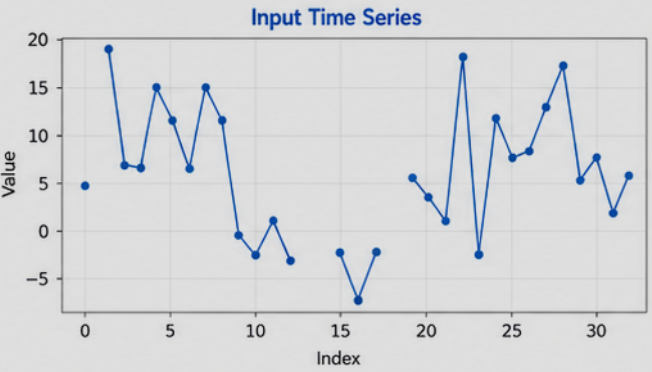}
\end{minipage}\hfill
\begin{minipage}[t]{\linewidth}\small 
  \textbf{Question Type:} TF
\end{minipage}
\end{tcolorbox}
\vspace{0.6em}
\begin{tcolorbox}[
  enhanced, breakable, title=Question \& Answer,
  colback=black!1, colframe=black!20, boxrule=0.5pt, title filled=false
]
\small
\medskip
\textbf{Question:} 
\\ Based on the given time series input, is the given input an irregular time series?
Respond ONLY with the letter of the correct choice (T or F).\\

Choices:
\begin{enumerate}[leftmargin=*, itemsep=2pt]
  \item[T:] True.
  \item[F:] False.
\end{enumerate}
\medskip
\textbf{Answer:} \textcolor{blue}{\textbf{T}}
\end{tcolorbox}
\end{tcolorbox}

\begin{tcolorbox}[
  enhanced, breakable, title= IRTS-ToolBench — Irregularity Cause Attribution, colback=grey, colframe=black!70, boxrule=0.8pt,
  colbacktitle=black!85, coltitle=white, fonttitle=\bfseries, title filled,
  sharp corners=all, arc=1.5mm, left=2mm, right=2mm, top=2mm, bottom=2mm
]
\begin{tcolorbox}[
  enhanced, breakable, title=Time Series Info ,
  colback=black!1, colframe=black!20, boxrule=0.5pt, title filled=false
]
\begin{minipage}[t]{1\linewidth}
  \includegraphics[height=6cm, width=\linewidth]{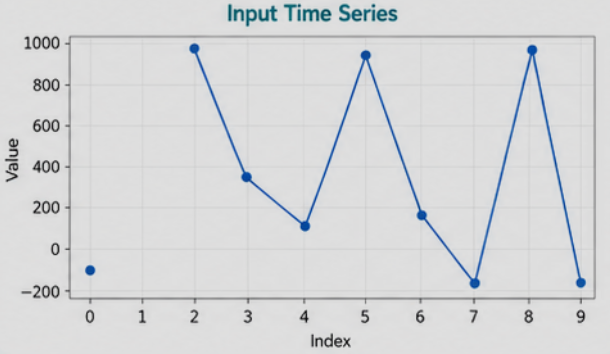}
\end{minipage}\hfill
\begin{minipage}[t]{\linewidth}\small 
  \textbf{meta\_info}: Human Activity Recognition wearable motion-sensor stream. A single sample is absent at the second timestamp, annotated as collection\_missing\_point / nan\_insertion\_point, attributed to temporary physical detachment or sensor buffering... \\

  \textbf{Question Type:} MC
\end{minipage}
\end{tcolorbox}
\vspace{0.6em}
\begin{tcolorbox}[
  enhanced, breakable, title=Question \& Answer,
  colback=black!1, colframe=black!20, boxrule=0.5pt, title filled=false
]
\small
\medskip
\textbf{Question:} 
\\ Identify the most likely irregularity category for this data collection mechanism.\\

Choices:
\begin{enumerate}[leftmargin=*, itemsep=2pt]
  \item[A:] Constraint-Based.
  \item[B:] Artifact-Based.
  \item[C:] Trigger-Based
\end{enumerate}
\medskip
\textbf{Answer:} \textcolor{blue}{\textbf{B}}
\end{tcolorbox}
\end{tcolorbox}

\begin{tcolorbox}[
  enhanced, breakable, title= IRTS-ToolBench — Missingness Reasoning, colback=grey, colframe=black!70, boxrule=0.8pt,
  colbacktitle=black!85, coltitle=white, fonttitle=\bfseries, title filled,
  sharp corners=all, arc=1.5mm, left=2mm, right=2mm, top=2mm, bottom=2mm
]
\begin{tcolorbox}[
  enhanced, breakable, title=Time Series Info ,
  colback=black!1, colframe=black!20, boxrule=0.5pt, title filled=false
]
\begin{minipage}[t]{1\linewidth}
  \includegraphics[height=6cm, width=\linewidth]{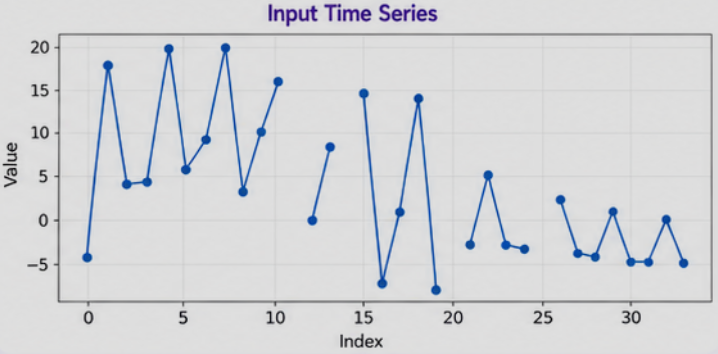}
\end{minipage}\hfill
\begin{minipage}[t]{\linewidth}\small 
  \textbf{meta\_info}: Short wearable-sensor signals capturing human movement. Partial observations arise from duty cycling, wireless packet loss, buffering delays, off-body periods, or app pauses... \\

  \textbf{Question Type:} MC
\end{minipage}
\end{tcolorbox}
\vspace{0.6em}
\begin{tcolorbox}[
  enhanced, breakable, title=Question \& Answer,
  colback=black!1, colframe=black!20, boxrule=0.5pt, title filled=false
]
\small
\medskip
\textbf{Question:} 
\\ Identify the most likely reason for the 2nd anomalous or missing observation.\\

Choices:
\begin{enumerate}[leftmargin=*, itemsep=2pt]
  \item[A:] Resampling bug at 14.0.
  \item[B:] Wrist Impact Saturation.
  \item[C:] Electromagnetic Interference.
  \item[D:] Sensor Buffer Dropped a Frame at 14.0.
\end{enumerate}
\medskip
\textbf{Answer:} \textcolor{blue}{\textbf{D}}
\end{tcolorbox}
\end{tcolorbox}

\end{document}